\documentclass[review]{elsarticle}

\usepackage{lineno,hyperref}
\modulolinenumbers[1]

\journal{arXiv}
\usepackage[utf8]{inputenc}
\usepackage{amsmath,amsthm,amsfonts}
\usepackage{enumitem}
\usepackage{graphicx}
\usepackage{color}
\usepackage{float}
\usepackage{fixltx2e} 
\usepackage{placeins}
\usepackage{todonotes}

\usepackage{setspace}
\graphicspath{ {./}}
\usepackage{geometry}
\newcounter{fig_num} 
\begin{document}
\begin{frontmatter}
\title{DeepAngle: Fast calculation of contact angles in tomography images using deep learning}

\author[address1]{Arash Rabbani \corref{mycorrespondingauthor}}
\cortext[mycorrespondingauthor] {Corresponding author}
\ead{a.rabbani@leeds.ac.uk | rabarash@gmail.com}
\author[address2]{Chenhao Sun}
\author[address3]{Masoud Babaei}
\author[address3]{Vahid J. Niasar}
\author[address4]{Ryan T. Armstrong}
\author[address4]{Peyman Mostaghimi}

\address[address1]{The University of Leeds, School of Computing, Leeds, UK}
\address[address2]{China University of Petroleum, Beijing, China}
\address[address3]{The University of Manchester, School of Chemical Engineering and Analytical Science, Manchester, UK}
\address[address4]{The University of New South Wales, School of Minerals and Energy Resources Engineering, Sydney, Australia}

\begin{abstract}
DeepAngle is a machine learning-based method to determine the contact angles of different phases in the tomography images of porous materials. Measurement of angles in 3--D needs to be done within the surface perpendicular to the angle planes, and it could become inaccurate when dealing with the discretized space of the image voxels. A computationally intensive solution is to correlate and vectorize all surfaces using an adaptable grid, and then measure the angles within the desired planes. On the contrary, the present study provides a rapid and low-cost technique powered by deep learning to estimate the interfacial angles directly from images. DeepAngle is tested on both synthetic and realistic images against direct measurement technique and found to improve the r-squared 5 to 16\%, while lowering the computational cost 20 times. This rapid method is especially applicable for processing large tomography data and time-resolved images, which is computationally intensive. The developed code and the dataset are available at an open repository on GitHub (\url{https://www.github.com/ArashRabbani/DeepAngle}).     
   
\end{abstract}

\end{frontmatter}


\section{Introduction}

The presence of two or more fluid phases within a porous material at the same time can create complex hydrostatic configurations \cite{wang2016understanding}. Depending on the chemistry of the solid walls, surface roughness \cite{ryan2008roughness}, environment temperature \cite{blake2019temperature}, pressure \cite{sarmadivaleh2015influence}, and fluid interfacial forces, many different arrangements occur as an outcome of interplay between the surface forces. Characterizing the behavior of multiphase fluids on solid surfaces has significant applications in the design of fuel cells, filters, condensers, and many chemical reactors. Additionally, such an arrangement of fluids in porous solids is present in many nature-made porous systems, from subsurface water resources to petroleum reservoirs, which highlights the importance of using quantitative methods to better understand the underlying processes. The contact angle at which the fluids touch the surface of solid walls is a significant parameter that describes the tendency of each fluid to cover the surface. Traditionally, it has been measured on smooth and flat surfaces based on experimental techniques, such as captive bubble and pendant drop methods. However, in-situ apparent contact angles in porous media present a wide distribution due to the complex pore topology, surface chemistry and roughness \cite{sun2020probing}. Therefore, a single angle measured on a flat surface cannot represent the wetting characteristics inside the porous medium.  

With recent advances of micro-computed tomography (micro-CT) imaging, it has become possible to track the interfaces of fluids while they interact with the surface of the porous materials. To date, there have been several previous attempts to measure the fluid contact angle directly from micro-CT images. Andrew et al. \cite{andrew2014pore} measured the in-situ apparent contact angles from manually selected raw micro-CT image set in a $CO_{2}$-brine-carbonate system. However, manual contact angle measurements may be prone to user bias. Consequently, automatic contact angle measurement methods that are based on curvature fitting have been developed by many previous works\cite{klise2016automated,lv2017situ,scanziani2017automatic,alratrout2017automatic,dalton2020contact,ibekwe2020automated,khanamiri2020contact}. These methods involved segmenting the image voxels of a 3-D system into respective phases, subsequently obtaining the three-phase contact points, and calculating the angle through the wetting phase between the solid surface and the tangent line of the fluid interface at each point. In contrast to measuring the microscopic contact angles along the three-phase contact line, Sun et al. \cite{sun2020probing,sun2020characterization,sun2022universal} developed a new method to characterize the wettability of the porous medium by capturing the fluid topology from micro-CT images based on a principle of integral geometry known as Gauss-Bonnet theorem (GBT) that links the topology of an object to its total curvature \cite{arakida2018light,allendoerfer1943gauss}. Measure curvatures using GBT can be normalized with respect to the contact line of a fluid cluster which makes in unbiased to the length of the contact line. In addition, GBT is less susceptible to partial-volume effects \cite{sun2020probing} and segmentation errors at the three-phase contact line when compared with the direct measurement of microscopic apparent contact angles. However, based on an unpublished pilot test by authors, this method still demands for virtually the same amount of computational resources as a direct measurement technique.   

Considering the high computational cost of the already available methods, especially when dealing with large-size tomography images, machine learning techniques can be used to speed up the process. A machine learning model can be directly trained on the voxelized tomography images to predict the contact angle without any need for vectorizing and smoothing the geometries, which is computationally intensive. Such a rapid approach can even make it possible to track the changes in the contact angle distribution in real time as the experiment is being performed. Image-based machine learning models have a long history of application in modeling and simulation of processes in porous material \cite{rabbani2021review,tahmasebi2020machine}. It has been shown that several common characteristics of porous media from absolute permeability \cite{kamrava2020linking,rabbani2019hybrid,rabbani2017estimation}, specific surface \cite{alqahtani2020machine}, and fracture morphology \cite{singh2021computer} to more complex features such as capillarity \cite{zhang2020accelerating,rabbani2020deepore} and gas adsorption \cite{rabbani2021image} can be estimated using artificial neural network with a competitive accuracy compared to numerical simulation techniques \cite{rabbani2020deepore}. 
In the present study, using the power of deep learning to bypass computationally expensive calculations, a rapid and novel method of contact angle estimation for fluid-solid interactions is developed. The results show that the proposed method is an accurate and computationally-efficient approach to characterize the wettability in porous media compared to the well-established direct measurement method of contact angles. 

\section{Methodology}
\label{sec:me}
In this section, the training and prediction stages of the proposed machine learning method is discussed. The final goal is to be able to measure the distribution of the interfacial contact angle from 3-D micro-CT segmented images in a rapid but still accurate manner.  

\subsection{Generating the training dataset}
\begin{figure}[H]
	\centering\includegraphics[page=\value{fig_num},width=.8\linewidth]{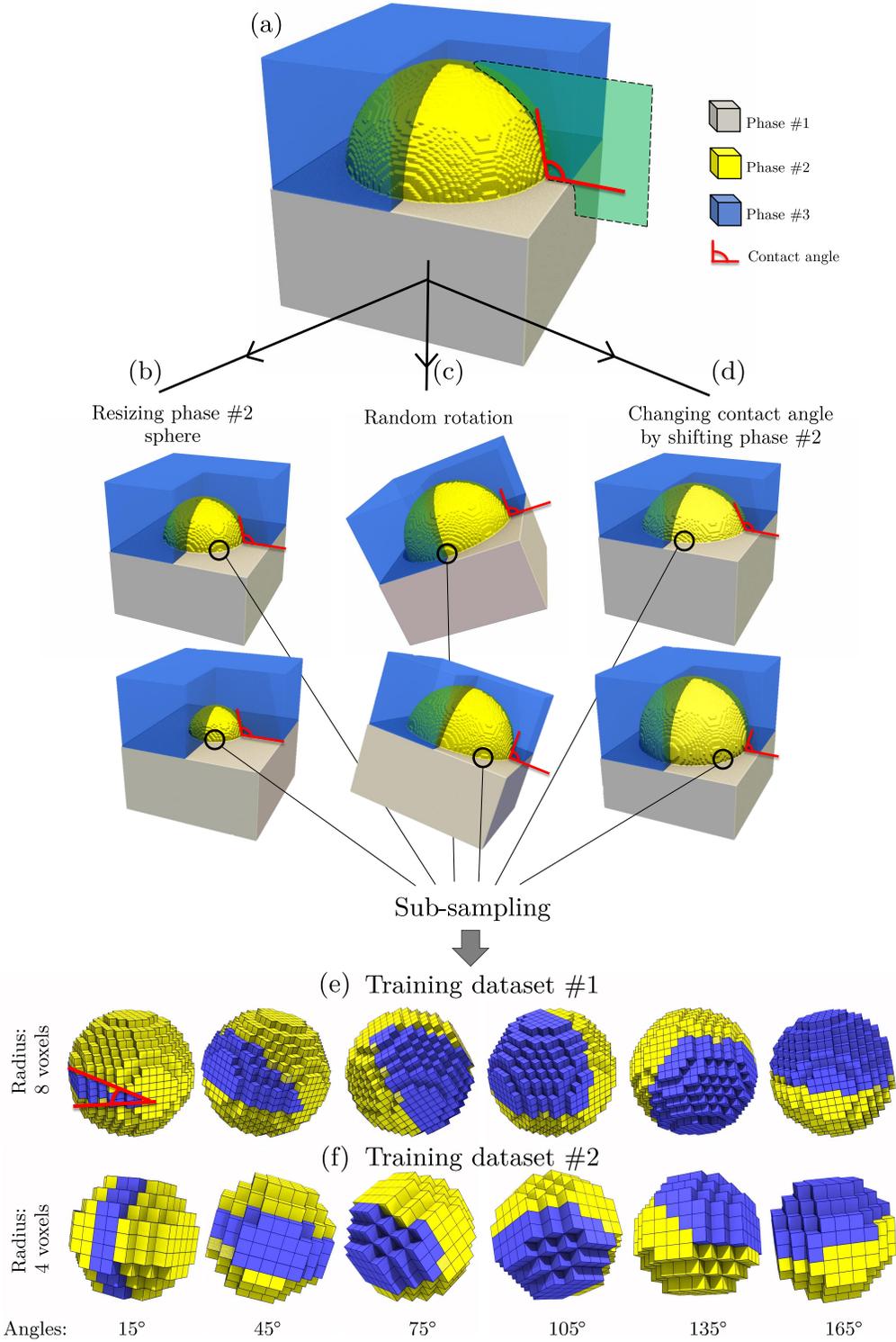} \stepcounter{fig_num}
	\caption{Preparation of the training dataset based on synthetic droplets of a fluid sitting on a flat surface in the presence of secondary fluid. a) 3D rendering of a droplet and the desired contact angle, b) augmentation of the dataset by resizing the sphere, c) augmentation of the dataset by rotating the voxelized geometry, d) increasing the number of training data points by vertical shifting of the sphere, e) sub-sampled training dataset \#1 with 8 voxels radius, f) sub-sampled training dataset \#2 with 4 voxels radius.}
	\label{fig:traindata}
\end{figure}

To train a deep learning model, we have generated a synthetic dataset of droplets placed on a flat surface surrounded by a second fluid phase, as illustrated in Fig. \ref{fig:traindata}. The droplet is actually a cap separated from a sphere that, by definition, has the same curvature radius in all surface points. To make a diversified dataset for training, we have voxelized the perfect geometry of the sphere into a grid space, which creates some roughness at the interfacial surface, as visible in Fig. \ref{fig:traindata}-a but helps to get closer to the real condition of the segmented tomography images. The synthetic geometry of the droplets has been diversified by random shifting (Fig. \ref{fig:traindata}-d) and resizing of the sphere (Fig. \ref{fig:traindata}-b) as well as random rotation of the whole geometry (Fig. \ref{fig:traindata}-c). In the next step small spherical sub-samples are obtained from the diversified images at the locations where three phases meet each other. These sub-samples are retrieved in two radii of 8 and 4 voxels, respectively, visible in Fig. \ref{fig:traindata}-e and Fig. \ref{fig:traindata}-f. It should be noted that the actual radii are 8.5 and 4.5 voxels if we include the spheres central voxel; however we use the nominal values of 8 and 4 to avoid complexity. Due to the limited resolution of the voxelized space, contact angles smaller than $10^{\circ}$  and larger than $170^{\circ}$ cannot be easily captured. Sub-samples are binarized into ones and zeros, with zeros indicating the surrounding fluid, and ones indicating the voxels within the solid or the droplet phase. The voxel values of each spherical sub-sample are flattened to form a 1-D array that acts as the input of the deep learning model. Considering that geometries are generated synthetically, the theoretical value of the contact angle ($\theta$) can be calculated by simple trigonometry.   
 
\begin{equation}
	\label{eq:ang}
	\theta=\pi/2-\arcsin(h/r)
\end{equation}

where $h$ is the height of the center of the sphere relative to the solid surface and takes negative values if the center is below the solid surface, and $r$ is the radius of the sphere with the same length unit as $h$. The contact angle obtained from Eq. \ref{eq:ang} is used as the output of the deep learning model.

\subsection{Deep learning models}
Size of the sub-samples is a critical parameter for accurate measurement of the angles. Very small sub-samples with radius lower than 2 voxels are not technically able to capture small angles, and very large sub-samples with radius greater than a typical droplet may include many irrelevant voxels that do not contribute to the intended contact angle. The scarcity of relevant data or the presence of irrelevant data can reduce the performance of the deep learning model. Through preliminary studies, we have found that small radius of sub-sample makes the model less precise but more accurate. Thus, to make a more robust model and decrease the uncertainties, we collect two sizes of sub-samples and use them for both training and prediction stages. The smaller size helps to improve the accuracy and larger size of the sub-samples enhance the precision of the model.

\begin{figure}[H]
	\centering\includegraphics[page=\value{fig_num},width=.8\linewidth]{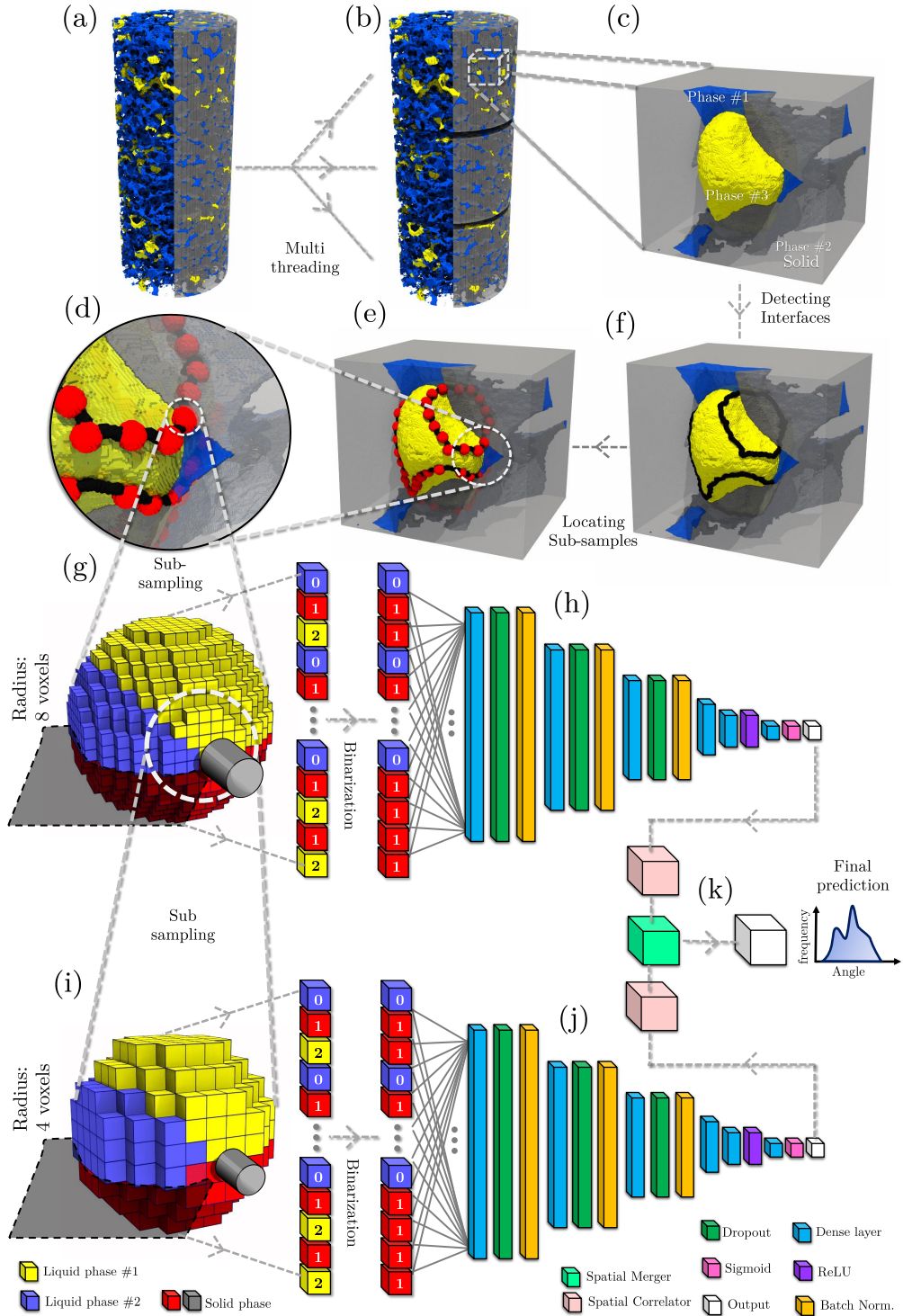} \stepcounter{fig_num}
	\caption{DeepAngle data workflow from segmented tomography image and inter-facial sub-sampling to prediction of the contact angles using two deep learning models via architecture \#2 and merging the results to obtain the final distribution of contact angle. a) segmented tomography image (raw data from \cite{singh2016imaging}), b) dividing the geometry into smaller sections to be analyzed by different computational threads, c) zoomed illustration of a non-wetting phase droplet (raw data from \cite{scanziani2017automatic}), d) detecting the three-phase interface line, e) taking spherical sub-samples centered at the voxels residing on the line, f) zoomed illustration of the sub-samples as red spheres, g) a sub-sample with the radius of 8 voxels, h) deep learning model architecture \#2 used to transform sub-section voxels into an angle, i) a sub-sample with the radius of 4 voxels, j) the secondary deep learning model used to transform sub-section voxels into a an additional estimation of the contact angle, and k) spatial correlator operator that averages each pairs of angles with a distance less than the sampling diameter and spatial merger which combines the predictions of the two deep learning models with different sub-sample radii.}
	\label{fig:workflow}
\end{figure}

Three deep learning model architectures have been devised to be tested on the training dataset. Architecture \#1 is a convolutional network with 17,621 and 17,845 trainable parameters, respectively, for the input radii of 4 and 8 voxels. This model contains two convolution stages with $3\times 3$ filter size and same-size padding each followed by a $2\times 2$ max-pooling and drop-out layers. Then a batch normalization layer is followed by three dense layers with rectified linear and sigmoid activation functions to generate the output.    
Architecture \#2 is a fully connected network with no convolutions. The first 3 initial dense layers with 128, 64, and 32 nodes are followed by drop-out and batch normalization layers. In addition, three dense layers with the size of 16, 4, and 1 nodes with rectified linear and sigmoid activation functions are used at the end of the network to generate the output value. This structure is depicted in Fig. \ref{fig:workflow}-g and i, as it has been found to be the optimal network architecture. The total trainable parameters of this model are 104,825 and 640,377, respectively, for inputs with radii of 4 and 8 voxels. Architecture \#3 is a shallower version of \#2 with 4 dense layers that contain 64, 16, 4, and 1 nodes followed by rectified linear and sigmoid activation functions on top of the last two dense layers. Total trainable parameters of this model are 47,961 and 315,737, respectively, for inputs with radii of 4 and 8 voxels. 
All designed deep learning models use the mean square error (MSE) as the loss function. Additionally, Adam optimizer with initial learning rate of $2\times10^{-4}$, exponential decay rate of 0.9 and decay steps of $2\times10^{4}$ are used for fitting the network weights and biases through error back-propagation technique. Deep learning models are implemented in the Python programming language using the Tensorflow package with Keras back-end. The developed code and the dataset are available at an open repository on GitHub \footnote{\url{https://www.github.com/ArashRabbani/DeepAngle}}. 

Prediction of the contact angle distribution begins with preprocessing of the segmented 3-D tomography image. Fig. \ref{fig:workflow} illustrates a schematic workflow that describes the angle measurement process. Initially, the geometry is divided into smaller sections and assigned to different computational threads for better use of the system resources through parallel computing (Fig. \ref{fig:workflow}-b). Now, let us focus on a small part of the geometry in which two fluid phases are in touch with the solid space (Fig. \ref{fig:workflow}-c) (raw data from \cite{singh2016imaging,scanziani2017automatic}). The interface in which all three phases meet each other can be found by searching the whole geometry with a $ 2^{3} $ voxels sliding window (Fig. \ref{fig:workflow}-f). A more computationally efficient approach for finding the interfacial line is to dilate the morphology of two phases by one voxel using an image dilation operation and find the intersection of the two dilated phases with the third phase. In the next step, several spherical sub-samples with the radius of 8 voxels and centers located on the interfacial line are randomly selected, which are shown with red spheres in Fig. \ref{fig:workflow}-e. Now let us focus on one of the spherical sub-samples (Fig. \ref{fig:workflow}-d). As it can be seen in Fig. \ref{fig:workflow}-g and i, initially all three phases are present in the sub-samples while by binarization we only keep the voxels that contain the surrounding fluid which in this example is represented by water. The reason is that traditionally, it is common to measure the contact angle within the water phase in the presence of oil and other non-aqueous fluids. Two deep learning models with fully connected dense layers with architecture \#2 are used to translate the binarized sub-samples into angle predictions (Fig. \ref{fig:workflow}-h and j). Each of the predicted angles is coupled with a coordinate that shows the center of the spherical sub-sample. As a result, a spatial distribution of the predicted contact angles is obtained. In case that two sub-samples overlap with each other, we replace their angle values with the average of both by assuming that the contact angle spatial function is continuous with no sudden jumps. The spatial correlator box in Fig. \ref{fig:workflow} represents the averaging process. Finally, in order to merge the predictions from the two deep neural networks, a local spatial operator is used to look into the vicinity of each 8-voxel sub-sample and shift the angle values to match the mean contact angles obtained from the 4-voxel model (spatial merger in Fig. \ref{fig:workflow}). As a result, a hybrid estimation of contact angles is obtained that contains the spatial coordinates for post-processing and interpretation purposes.

\subsection{Direct measurement method}
To test the performance of the developed deep learning approach for the prediction of contact angles, we have generated a second synthetic dataset with concave and convex solid surfaces to add more challenge. In addition to the DeepAngle, direct measurement approach that has been previously presented in the literature is used to demonstrate a comparison. In direct measurement approach, microscopic contact angles are measured along the three-phase contact line by using the 3D local method \cite{alratrout2017automatic}. The voxels of the micro-CT image were first segmented into respective phases. The segmented images are then smoothed by the Gaussian smoothing method to eliminate noise artifacts while preserving the internal structure of the porous material. Additional smoothing is applied to the fluid interface to obtain a constant curvature and fluid topology. This process is followed by identification of the three-phase contact points. As a result, two normal vectors are placed at each contact point, and the dot product of these two vectors is used to measure the apparent contact angle along the contact line. 
 
\section{Results and Discussions}
In this section, we initially discuss different architectures of deep learning models to predict contact angles. Then, using the model with the best performance, a comparison is presented to evaluate the DeepAngle results with other competing methods in the literature. This comparison has been performed using realistic and synthetic 3-D data. Finally, a time-resolved image dataset of fluid flow through porous media is used to demonstrate an application of the present method in the interpretation of the phenomena observed in experiments.    

\subsection{Deep learning model architecture}
\begin{figure}[H]
	\centering\includegraphics[page=\value{fig_num},width=.8\linewidth]{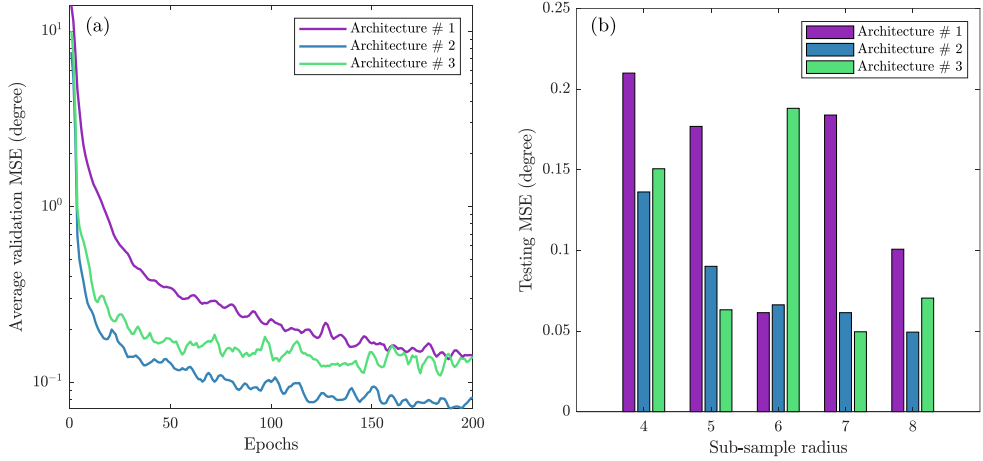} \stepcounter{fig_num}
	\caption{Comparison of the deep learning model architectures in terms of validation and testing loss measured by mean square error (MSE), a) average validation loss during 200 epochs of training on datasets over different sub-sample radius, b) comparing testing MSE for different architectures and sub-sample radius.}
	\label{fig:models}
\end{figure}

Two synthetic datasets of contact angle images have been generated using the approach presented in Fig. \ref{fig:traindata}. Each of the datasets contains 10000 3-D images with contact angles randomly distributed between 5 and 175 degrees. 80\%, 10\%, and 10\% of the data have been used for training, validation, and testing of the deep learning models. Based on the validation and test results, the model architecture \#2 was found to be more suitable for the prediction of contact angles. Architecture \#2 achieved a lower validation loss during the training process after 200 epochs compared to the other models (Fig. \ref{fig:models}-a). In addition, MSE of this model gradually decreases versus sub-sample radius from 0.14 to 0.05 degrees, while other models do not show this stable trend (Fig. \ref{fig:models}-b). It can be concluded that architecture \#2 is more suitable for image-based prediction of 3-D contact angles.  

\subsection{Comparison on synthetic data}
\begin{figure}[H]
	\centering\includegraphics[page=\value{fig_num},width=.8\linewidth]{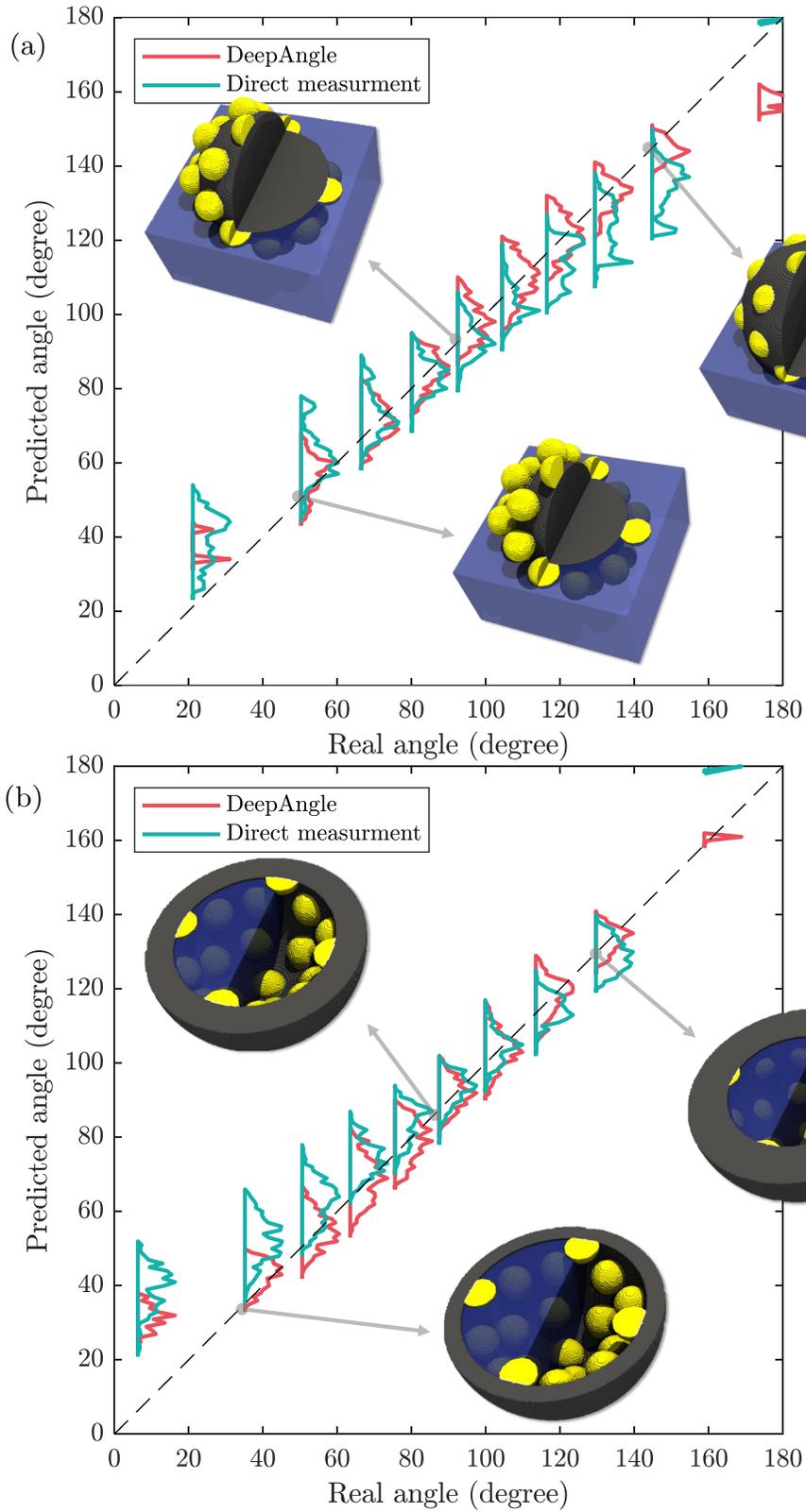} \stepcounter{fig_num}
	\caption{Comparison of the DeepAngle predictions and direct measurement, a) prediction of a range of contact angles on a convex surface, b) prediction of a range of contact angles on a concave surface. Gray material indicates the solid surface, blue is the wetting fluid an yellow denotes the non-wetting fluid. Solid spheres as well as other phases are cut for better visualization of the contact angles. }
	\label{fig:concave}
\end{figure}

\begin{figure}[H]
	\centering\includegraphics[page=\value{fig_num},width=.8\linewidth]{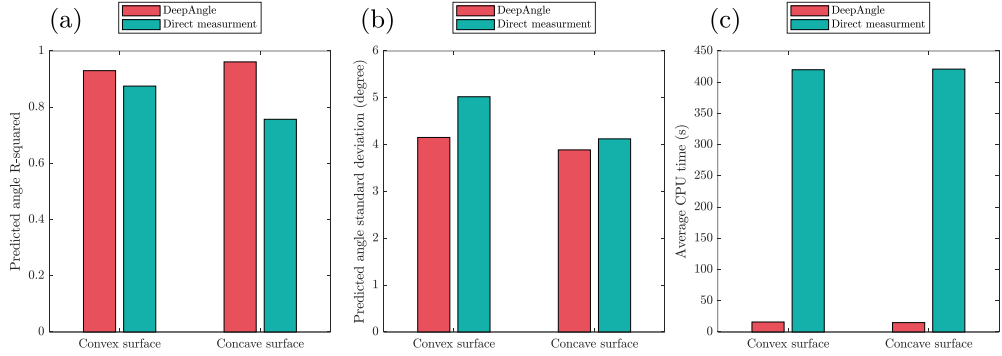} \stepcounter{fig_num}
	\caption{Comparison of the DeepAngle predictions and direct measurement method in terms of a) R-squared in prediction of the contact angles, b) standard deviation of the predicted angles, and c) approximated computational time for convex and concave geometries with the size $400^3$ voxels.}
	\label{fig:concave2}
\end{figure}

To evaluate the performance of the developed model, further contact angle measurements are performed on a synthetic dataset of droplets that sit on the inner or outer surface of a sphere (Fig. \ref{fig:concave}). In each of the 3-D volumetric images, 24 spherical droplets are located at equally distanced points on the surface of a sphere with a radius five times larger than the droplets. By modifying the distance between the center of spheres, the contact angle can be controlled, as visualized in Fig. \ref{fig:traindata}. Considering that DeepAngle has been trained on a dataset with flat solid surfaces, testing the effect of surface curvature helps to ensure that the model is not biased to the training condition. In addition, direct measurement method is used to present a comparative evaluation. Fig. \ref{fig:concave}-a and b illustrate the distribution of the predicted contact angles on convex and concave surfaces. By approaching two ends of the contact angle spectrum, either 0 or 180, the uncertainty of both methods increases, which is justifiable considering the limited image resolution for capturing very small or very large angles. Additionally, the coefficient of determination ($R^2$) and the standard deviation of the predicted angles are calculated to serve as comparison criteria (Figs. \ref{fig:concave2}-a and b). $R^2$ of the DeepAngle predictions are 0.93 and 0.96 respectively for convex and concave surfaces while direct measurement led to $R^2$ of 0.88 and 0.76. In addition, the standard deviation of the DeepAngle predictions are on average around 10\% lower than the direct measurement technique which indicates a higher precision. In addition to a higher accuracy and precision, DeepAngle is computationally more efficient. The approximate computational times of the two methods are measured using a single-core Intel processor with 3.0 GHz frequency to process 20 volumetric images with $400^3$ voxels. As it can be seen in Fig. \ref{fig:concave2}-c, the average computational cost of the DeepAngle is around 20 times lower than the direct measurement method.

\subsection{Comparison on realistic data}
In this section, a comparison is presented between the predicted contact angles in realistic data. The data are originally produced and analyzed by Alhammadi \textit{et al.} \cite{alhammadi2017situ} and the images are publicly available on the Digital Rocks Portal \cite{prodanovic2015digital}. In this dataset, three tomography images of porous sandstone rock samples are segmented after primary drainage with crude oil in the presence of water. For information regarding the rock and fluid properties as well as the aging and experimental protocol, readers are refer to the appendices of Alhammadi \textit{et al.} \cite{alhammadi2017situ}. Images are in the size of $976\times1014\times601$ voxels with mean contact angles of $77^{\circ} \pm21^{\circ}$, $104^{\circ}\pm26^{\circ}$, and $94^{\circ}\pm24^{\circ}$, respectively, for samples \#1 to \#3 measured using the method presented by Alhammadi \textit{et al.} \cite{alhammadi2017situ}. They have employed surface fitting to convert the voxelized geometry of fluid droplets and solid space into a vectorized shape in which the measurement of angles is more robust. However, the high computational cost of this direct approach is a bottleneck, especially when analyzing large-size images. 

\begin{figure}[H]
	\centering\includegraphics[page=\value{fig_num},width=.8\linewidth]{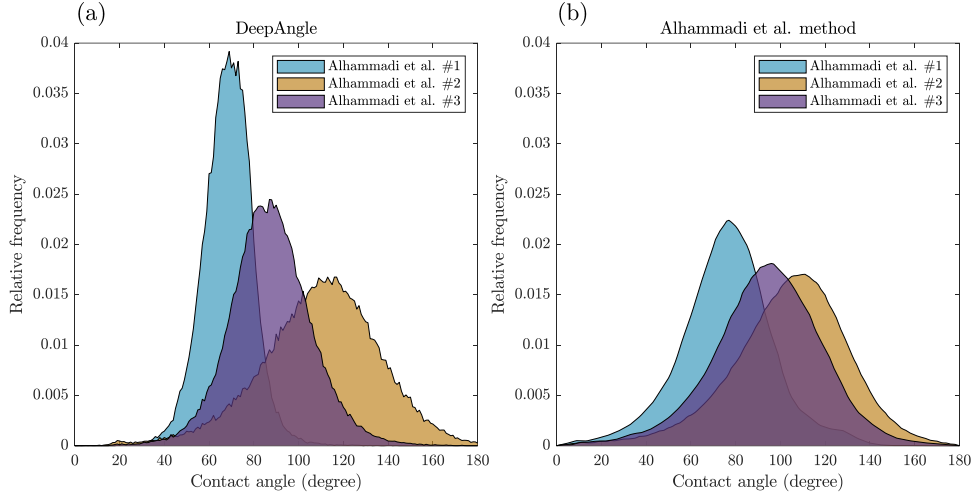} \stepcounter{fig_num}
	\caption{Comparison of the contact angle predictions by  DeepAngle and Alhammadi \textit{et al.} method \cite{alhammadi2017situ} for three porous rock samples saturated with water and crude oil.}
	\label{fig:alhammadi}
\end{figure}

This dataset of realistic images has been analyzed by DeepAngle to evaluate the similarity. The mean contact angles obtained are $68^{\circ} \pm11^{\circ}$, $109^{\circ}\pm26^{\circ}$, and $87^{\circ}\pm18^{\circ}$, respectively, for samples \#1 to \#3. The average contact angles have around 8\% differences, while the standard deviation of the DeepAngle predictions is around 31\% lower than the Alhammadi \textit{et al.} method \cite{alhammadi2017situ}. Both contact angle distributions are visualized in Fig. \ref{fig:alhammadi}). The higher standard deviation of the angles in the Alhammadi \textit{et al.} method can be justified by considering that the surface smoothing process in their approach may have flatten the contact angle distributions to some extent.      

\subsection{Time-resolved angle measurement}

\begin{figure}[H]
	\centering\includegraphics[page=\value{fig_num},width=.8\linewidth]{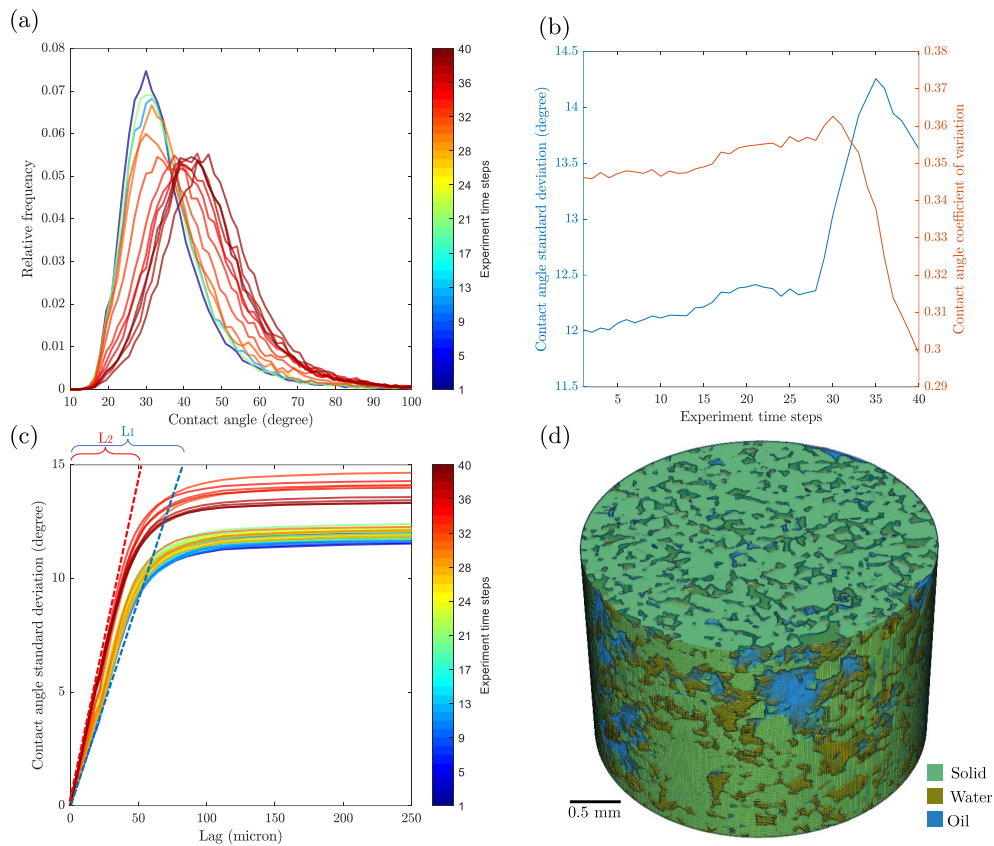} \stepcounter{fig_num}
	\caption{Spatio-temporal changes of contact angle distribution during an imbibition experiment performed by Rucker \textit{et al.} \cite{rucker2015connected}, a) contact angle distributions for 40 time-steps of the experiment, b) changes in the standard deviation and coefficient of variation of the contact angles, c) spatial changes in the standard deviation of contact angles, $L_1$ is equivalent to the correlation length at the beginning and $L_2$ is equivalent to the correlation length at the end of the experiment, d) volumetric visualization of the solid, water (wetting) and oil (non-wetting) phases at the beginning of the experiment. }
	\label{fig:dyn1}
\end{figure}

Measurement of the contact angle in a dynamic time-resolved experiment can provide new insights into the interpretation of the results. Considering the low computational time of the DeepAngle approach, which is below 20 seconds for an image with $400^3$ voxels using a common desktop computer, it is a promising method to tackle big data time-resolved experiments that come with terabytes of tomography data. To demonstrate an example, we have used a time series of 3-D volumetric data published by Rucker \textit{et al.} \cite{rucker2015connected} to analyze the contact angle spatio-temporal changes during an imbibition experiment. During imbibition, a wetting fluid is injected into a porous material that is partially saturated with a secondary non-wetting fluid. The wetting phase, which tends to occupy tight corners and small pores as a result of the capillary forces, does form a continuous film that surrounds the non-wetting phase. As the saturation of the wetting phase increases, the film swells, and it may snap-off some connections between the continuous phase of the non-wetting fluid and separate it to isolated ganglia. This change in the flow regime is an interesting topic to study due to its wide application in subsurface energy sources and groundwater remediation \cite{pak2020pore,aminnaji2019effects}. In this section, we have analyzed 40 time-steps of the imbibition experiment performed by Rucker \textit{et al.} \cite{rucker2015connected} and the contact angle changes during the experiment are visualized in Fig. \ref{fig:dyn1}. In this figure, as the flow regime changes from film swelling to ganglia mobilization, the distribution of the contact angle becomes more flat and diversified (Fig. \ref{fig:dyn1}-a). This observation is more clearly visible in Fig. \ref{fig:dyn1}-b with an increase in the angle standard deviation. However, the increase in average contact angle is more significant compared to the increase in standard deviation that led to the overall decrease in the coefficient of variation, which is obtained by dividing the standard deviation by average. Additionally, the standard deviation of the contact angles can be measured on different length scales. Fig. \ref{fig:dyn1}-c visualizes the standard deviation of the angle based on an average lag distance between the sampled data points. In this figure, the intersection length of the tangent lines and a constant standard deviation line can be interpreted as the correlation length ($L_1$ and $L_2$). As can be seen, the contact angle becomes more spatially diversified when moving toward the ganglia flow regime, which is reasonable considering the independence of the isolated ganglia in capillary force balance. Additionally, it can be interpreted that in ganglia flow regime, the leading edge of the gangalion has an advancing angle due to drainage nature of the process while the tail is experiencing imbibition which led to observe a receding angle. Such difference can create a more diversified set of angles compared to the film swelling regime which only represent drainage with advancing angles.   

In addition, spatial distribution of the contact angles can be used for characterization of the porous material wettability. Solid structures composed of materials with different surface properties may show different wetting behavior. In Fig. \ref{fig:dyn1} a section of the porous medium studied by Rucker \textit{et al.} \cite{rucker2015connected} is visualized along with the spatial contract angle distribution obtained from the first time step. The interpolation of the contact angles from a point cloud representation (Fig. \ref{fig:dyn1}-b, left) to a volumetric representation (Fig. \ref{fig:dyn1}-b, right) is implemented by morphological image dilation. Experimental in-situ characterization of wettability can be useful for making more accurate simulation models, which enables researchers to simulate experiments with higher fidelity. For example, volumetric interpolated contact angles can be used to realistically initialize two-phase computational fluid dynamic models and improve their prediction capability. Such numerical modeling applications are suggested to be investigated in future studies.    

\begin{figure}[H]
	\centering\includegraphics[page=\value{fig_num},width=.7\linewidth]{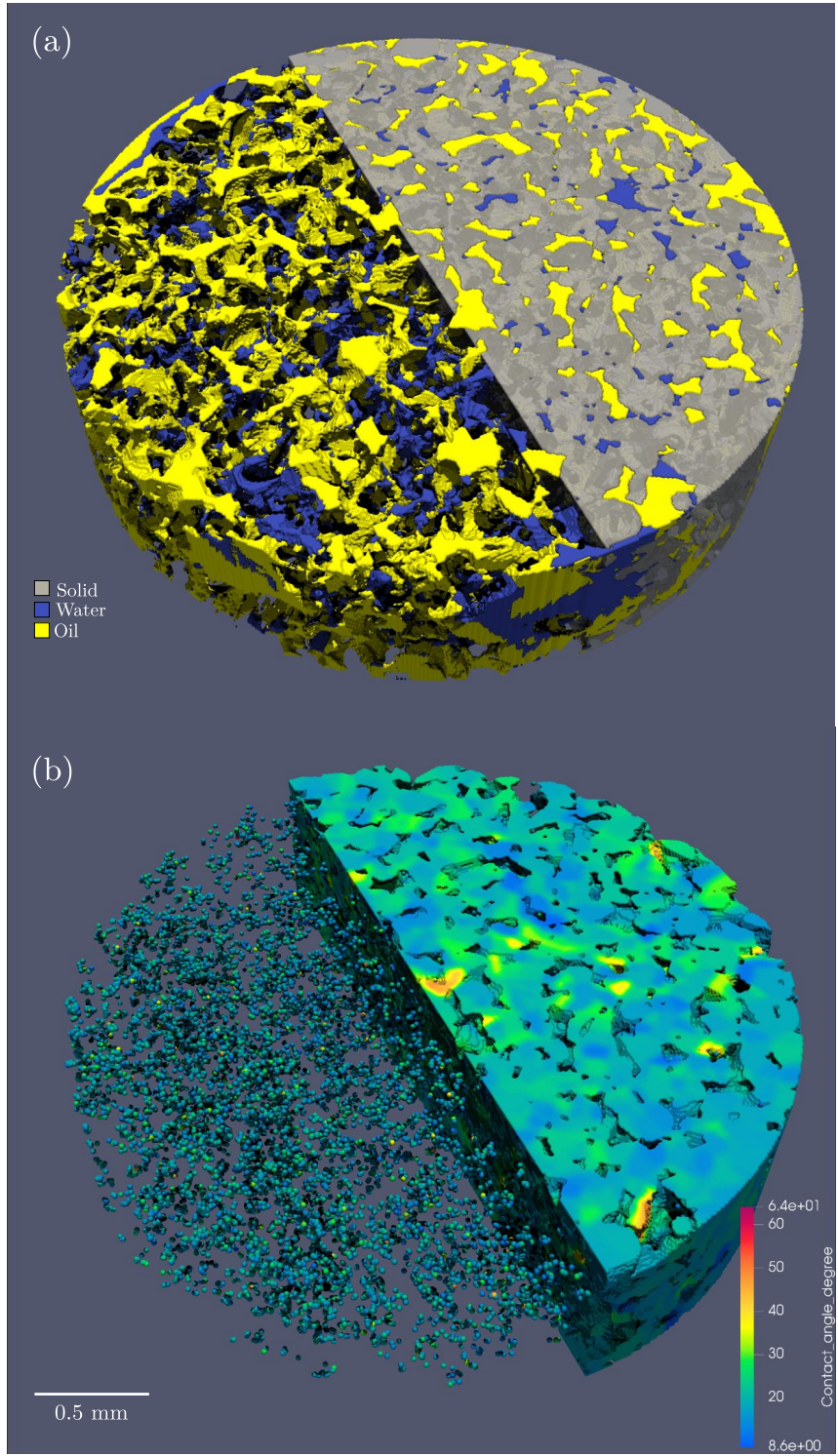} \stepcounter{fig_num}
	\caption{a) 3-D illustration of the fluids within the top 20\% of the sample at the first time-step, b) point-cloud (left) and volumetric (right) representation of the interpolated contact angle for the same geometry.}
	\label{fig:dyn2}
\end{figure}

\section{Conclusions}

\label{sec:con}
In this study, we have developed and presented a deep learning method (DeepAngle) to measure 3--D contact angles in tomography images in which a porous material is saturated with two immiscible fluids. The common practice in the literature is to vectorize the voxelized geometries by fitting curved surfaces and consequently measure the angle between the fitted surfaces at three-phase interfacal points. In contrast, DeepAngle skips the vectorization step and estimates the angles by direct analysis of voxelized image. Such advantage can reduce the computational cost up to 20 times and still improving the precision and accuracy compared to the direct measurement technique. This rapid method is especially applicable for processing large tomography data and time-resolved images, which are computationally intensive. Finally, a summary of the findings of this study can be drawn as follows. 
\begin{itemize}
	\item  Three different deep learning models have been tested for contact angle measurement, and architecture \#2 which had fully connected dense layers was the best in terms of better prediction of test data and lower validation loss. This structure is a fully connected dense neural network with no convolution layers and maintained a low testing MSE for 5 tested sizes of sub-samples.     
	\item  DeepAngle is tested against direct measurement technique to measure the contact angle on curved solid surfaces. The determination coefficient ($R^2$) of the DeepAngle are 0.93 and 0.96, respectively, for convex and concave surfaces while direct measurement led to ($R^2$) of 0.88 and 0.76.  
	\item  The distribution of contact angles in three different realistic porous samples is measured and compared with the published results in the literature. The average contact angles have around 8\% difference, while the standard deviation of the DeepAngle predictions was around 31\% lower than the previous method.   
	\item A spatio-temporal analysis of contact angle on a two-phase dynamic fluid flow experiment is implemented to demonstrate the application of the developed method in tacking a large volume of data. Based on the analysis presented, it has been observed that during the imbibition experiment, as the flow regime changes from film swelling to ganglion--dominant, contact angles become more spatially diversified.

\end{itemize}
\section*{Computer code availability}
The developed code and the dataset are available at an open repository on GitHub (\url{https://www.github.com/ArashRabbani/DeepAngle}). 

\section*{Authorship statement}
AR has contributed in developing the main idea, computer programming and writing. CS has contributed in computer programming and writing, MB, VN, RA, and PM have contributed in writing and reviewing the paper. 
\newpage
\bibliography{ref}

\end{document}